\documentclass[preprint,11pt]{elsarticle}

\usepackage[T1]{fontenc}
\usepackage{times}
\usepackage[margin=1.15in]{geometry}
\usepackage{amssymb}
\usepackage{amsmath}
\usepackage{graphicx}
\usepackage{hyperref}
\usepackage{booktabs}
\usepackage{multirow}
\usepackage{xcolor}
\usepackage{url}
\usepackage{siunitx}
\usepackage{subcaption}
\usepackage{tabularx}
\usepackage{array}
\usepackage{enumitem}
\usepackage[skins]{tcolorbox}
\usepackage{tikz}
\usetikzlibrary{arrows.meta}

\setlist[itemize]{leftmargin=16pt, itemsep=2pt, topsep=2pt}
\setlist[enumerate]{leftmargin=16pt, itemsep=2pt, topsep=2pt}

\newcommand{\cmk}{{\color{green!55!black}\checkmark}}
\newcommand{\dsh}{{\color{gray}--}}

\newcommand{\Paragraph}[1]{\medskip\noindent\textbf{#1}}

\definecolor{objboxbg}{RGB}{247,240,208}
\definecolor{objboxframe}{RGB}{120,110,80}
\newtcolorbox{objbox}{
  colback=objboxbg!35,
  colframe=objboxframe,
  arc=2pt,
  boxrule=0.4pt,
  left=8pt,right=8pt,top=6pt,bottom=6pt,
}

\usepackage{listings}
\journal{HardwareX}

\begin{document}

\begin{frontmatter}

\title{NeoRacer: An Open, Standardized 1:12 Scale Autonomous Race Car for Benchmarking and Education }

\author[neo]{Koneshka Bandyopadhyay\corref{cor1}}
\ead{kb@neobotics.org}
\cortext[cor1]{Corresponding author}

\author[neo]{Ansh Mehta}
\ead{anshm@neobotics.org}

\author[neo]{Bassel El Mabsout}
\ead{bmabsout@neobotics.org}

\author[bu]{Renato Mancuso}
\ead{rmancuso@bu.edu}

\affiliation[neo]{organization={Neobotics Foundation, Inc.},
  city={Cambridge},
  state={MA},
  country={USA}}

\affiliation[bu]{organization={Boston University},
  city={Boston},
  state={MA},
  country={USA}}

\begin{abstract}
While many existing fields in scientific domains depend on standard benchmarks for tackling research problems (e.g. Machine Learning), easing the burden of reviewing and reproducing various methods within the countless manuscripts, autonomous systems research has a significant lack of open, accepted, and prevalent hardware of use. Where standardization has appeared, the field has flourished. The issue presents itself poignantly in autonomous racing, as groups often build custom solutions, or buy niche and expensive research vehicles. Making applied control research, robotics research, and education untranslateable and difficult to compare. Furthermore, the expensive nature of such vehicles make exploration unavailable outside established labs with very significant resources. The nearest educational robots are affordable, but heavily underpowered and uninspiring.
Thus, this work contributes the NeoRacer, an open-source 1:12 scale autonomous racing platform designed to bridge this gap. Built around an NVIDIA Jetson Orin Nano (67~TOPS), a 270\textdegree{} LiDAR, a 120~fps global shutter camera, and a 9-axis IMU, NeoRacer ships pre-assembled and ready to use for \$2,699, delivering over 3$\times$ the compute of comparable platforms at less than half the cost of the nearest pre-assembled alternative. NeoRacer is co-developed by the Neobotics Foundation and Seeed Studio and manufactured by Seeed Studio, pairing an open hardware and software design with scaled, repeatable production. The platform is designed to be modular and extensible, providing a standardized benchmarking environment for autonomous racing algorithms across institutions. We describe the hardware and software architecture, the design decisions informed by two pilot deployments (MIT IAP, 15 students; BU CPS Lab, 10 students), and the cost-performance tradeoffs that guided the platform's development. NeoRacer hardware is licensed under CERN-OHL-S v2 and software under GPLv3, with all design files, firmware, and ROS2 software packages publicly accessible.
\end{abstract}


\end{frontmatter}

\section{Introduction}
\label{sec:introduction}

Autonomous vehicles represent one of the most interdisciplinary engineering challenges of the 21st century, requiring expertise in perception, planning, control, machine learning, and real-time systems. Yet the educational infrastructure for teaching these skills at the university and K--12 level remains fragmented. Existing platforms fall into two categories: affordable kits with limited compute that cannot run modern perception algorithms (e.g., DuckieBot~\cite{duckietown2017}, PiCar-X~\cite{picarx}), and research-grade platforms costing over \$4,000 that are inaccessible to most educational budgets (e.g., F1Tenth~\cite{f1tenth2020}, MuSHR~\cite{mushr2019}).

This gap has three consequences. First, students at resource-constrained institutions are excluded from hands-on AV education. Second, institutions that can afford research platforms must invest significant time in assembly, driver configuration, and access to specialized fabrication equipment (laser cutters, 3D printers) before any teaching can begin. Third, the absence of a standardized, affordable platform prevents meaningful cross-institutional benchmarking of autonomous racing algorithms and inhibits the formation of a broad competitive ecosystem analogous to FIRST Robotics~\cite{first2000} or VEX Robotics~\cite{vex}.

We present NeoRacer, a 1:12 scale autonomous racing platform co-developed by the Neobotics Foundation and Seeed Studio and manufactured by Seeed Studio, designed to address all three problems simultaneously. NeoRacer is:

\begin{itemize}
  \item \textbf{Capable:} 67~TOPS of AI compute (NVIDIA Jetson Orin Nano), 270\textdegree{} LiDAR, 120~fps global shutter camera, 9-axis IMU, and encoder feedback. This is sufficient to run SLAM, deep learning inference, and a real-time control loop concurrently on a single onboard compute module.
  \item \textbf{Affordable:} \$2,700 pre-assembled, compared to \$6,000 for the nearest pre-assembled alternative (RACECAR/J F1Tenth~\cite{racecarj}).
  \item \textbf{Pre-assembled and modular:} Ships ready to use with all drivers pre-configured, eliminating the need for extra specialized equipment or multi-vendor part sourcing. Modular hardware and software design enables institutions to adapt the platform to their own curricula and research needs, and replacement parts are available from a single source when components are damaged.
  \item \textbf{A benchmarking standard:} Standardized, spec-locked hardware ensures that performance differences across teams and institutions are attributable to algorithms, not hardware advantages.
  \item \textbf{Competition-ready:} Fully compatible with the F1Tenth autonomous racing ecosystem and competition rules.
  \item \textbf{Open:} Hardware documentation licensed under CERN-OHL-S v2, software under GPLv3. Schematics, electrical and mechanical documentation, firmware, and all software sources are publicly available; PCB fabrication files are not yet released, and the next hardware revision is being developed to be fully open-hardware (\S\ref{sec:licensing}).
\end{itemize}

\begin{figure}[t]
\centering
\includegraphics[width=0.95\linewidth]{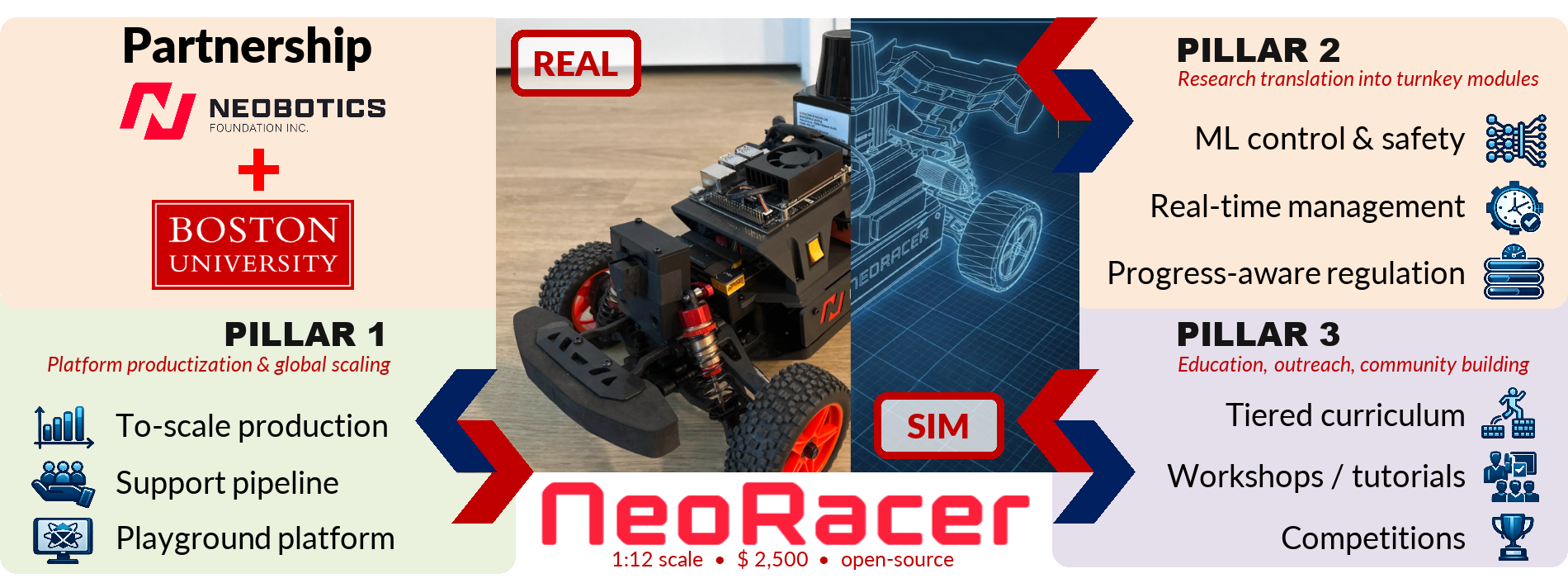}
\caption{Operating model for the NeoRacer platform, organized around three activity pillars: platform productization and scaling, research translation, and education and outreach. The platform is produced and maintained by the Neobotics Foundation and used as a research substrate by the Boston University Cyber-Physical Systems Lab. Each link is bidirectional: field feedback from deployments, research outputs, and educational use feeds continuously back into platform development. The pricing represents the pre-order price of \$2,500.}
\label{fig:operating_model}
\end{figure}

In this paper, we describe the hardware architecture (\S\ref{sec:hardware}), the release identity and manufacturing traceability scheme (\S\ref{sec:traceability}), the software stack (\S\ref{sec:software}), the research directions the platform is intended to enable (\S\ref{sec:research}), a reference curriculum design (\S\ref{sec:curriculum}), and cost analysis (\S\ref{sec:cost}). We report results from a pilot deployment at MIT (\S\ref{sec:evaluation}) and discuss the platform's role in enabling a global autonomous racing competition league (\S\ref{sec:competition}). Figure~\ref{fig:operating_model} summarizes the operating model around the platform and the feedback loops between production, research, and educational use.

\section{Related Work}
\label{sec:related}

\subsection{Small-Scale Autonomous Racing Platforms}

In response to the lack of autonomous racing benchmarks, various groups have built platforms for tackling the aforementioned problems:

\textbf{the F1Tenth project~\cite{f1tenth2020}}: established the 1:10 scale autonomous racing format as a viable educational and research testbed. Built around an NVIDIA Jetson Xavier NX (21~TOPS) with a Hokuyo UST-10LX LiDAR (\$1,200~\cite{hokuyoust10lx}), the F1Tenth platform supports SLAM, path planning, and competitive head-to-head racing. However, its official bill of materials~\cite{f1tenthbom} totals approximately \$3,800 in parts, requires laser cutters and 3D printers for chassis fabrication, and demands integration across components sourced from over a dozen vendors. This multi-vendor dependency also complicates repairs: replacing broken parts during high-speed racing requires re-sourcing from individual suppliers and potentially re-fabricating custom components, often taking platforms out of service for weeks. The only pre-assembled option, the RACECAR/J F1Tenth~\cite{racecarj}, is priced at \$6,000. Additonally, we found competitive teams frequently upgrade to higher-performance LiDARs such as the Hokuyo UTM-30LX-EW (\$4,375~\cite{hokuyoutm30lx}), and some now use Jetson Orin NX (100+~TOPS) compute modules, further increasing the cost and discouraging fair competition.

\textbf{The MIT RACECAR (Rapid Autonomous Complex Environment Competing Ackermann-steering Robot)~\cite{mitracecar2017} platform}: pioneered the use of scaled autonomous vehicles in university education, forming the basis for MIT's Beaver Works Summer Institute (BWSI) program. While influential, the original RACECAR hardware has become dated, and the RACECAR Neo successor~\cite{racecarneosite} uses a Raspberry Pi 4 supplemented by a Google Coral Edge TPU (4~TOPS combined). However, the Coral accelerator has become increasingly difficult to source due to chip shortages, and the platform's compute remains insufficient for modern perception pipelines that require simultaneous SLAM and deep learning inference.

\textbf{MuSHR (Multi-agent System for non-Holonomic Racing)~\cite{mushr2019}:} from the University of Washington forked and older version of the racecar neo, and provides a modern open-source Ackermann-steered platform built with a focus on multi-agent scenarios. Like the RACECAR NEO, its cost availability  present barriers to classroom-scale adoption.

\subsection{Low-Cost Educational Robots}

At the other end of the spectrum, platforms like DuckieBot~\cite{duckietown2017} (\$1,098 pre-assembled; from \$429 as a DIY kit), PiCar-X~\cite{picarx} (\$90 kit, compute sold separately), and AWS DeepRacer~\cite{deepracer} (\$399, now discontinued as of December 2025) offer lower entry points but lack the compute necessary for real-time deep learning inference alongside perception. The DuckieBot uses a Jetson Nano (0.5~TOPS) and includes wheel encoders but no LiDAR. AWS DeepRacer required cloud connectivity and incurred ongoing AWS training costs, creating a dependency that was problematic for classroom use.

TurtleBot4~\cite{turtlebot4} (\$1,850) provides a ROS2-native platform with an RPLIDAR A1M8, wheel encoders, and out-of-the-box SLAM support. However, it uses differential drive rather than Ackermann steering, has less than 1~TOPS of compute (Raspberry Pi 4), and is designed for indoor navigation rather than high-speed autonomous racing, which limits its utility for teaching vehicle dynamics, control theory, and racing algorithms.

ROSbot 3 PRO~\cite{rosbot3pro} (\$3,800) from Husarion offers a ROS2-native differential drive platform with LiDAR, encoders, and SLAM support, but uses a Raspberry Pi 5 with less than 1~TOPS of AI compute and is not designed for autonomous racing.

\subsection{Positioning NeoRacer}

Table~\ref{tab:comparison} summarizes the competitive landscape. While platforms like TurtleBot4 offer LiDAR and SLAM at a lower price point, they use differential drive and lack the compute necessary for real-time deep learning inference alongside SLAM. F1Tenth provides comparable sensor suites but costs \$6,000 pre-assembled while offering less than a third of NeoRacer's compute. NeoRacer is the only pre-assembled Ackermann-steered platform under \$3,000 that combines research-grade compute (67~TOPS) with a full perception suite. Critically, because all NeoRacer units ship with identical hardware specifications, the platform provides a standardized benchmarking environment where algorithmic innovations can be compared directly across institutions.

\begin{table}[h!]
\centering
\caption{Representative platform comparison showing where NeoRacer sits between classroom kits and research-grade autonomous platforms.}
\label{tab:comparison}
\renewcommand{\arraystretch}{1.18}
\setlength{\tabcolsep}{2pt}
\scriptsize
\begin{tabularx}{\linewidth}{@{}l *{7}{>{\centering\arraybackslash}X}@{}}
\toprule
 & \textbf{NeoRacer} & \textbf{DuckieBot}~\cite{duckietown2017} & \textbf{RACECAR Neo}~\cite{racecarneosite} & \textbf{TurtleBot~4}~\cite{turtlebot4} & \textbf{XBot~4WD}~\cite{xbot4wd} & \textbf{F1Tenth}~\cite{f1tenth2020} & \textbf{Jackal}~\cite{jackal} \\
\midrule
LiDAR              & 30\,Hz (25\,m) & \dsh{}      & 12\,Hz (12\,m) & 10\,Hz (12\,m) & 12\,Hz (30\,m) & 40\,Hz (30\,m) & configurable \\
Compute            & Orin Nano      & Jetson Nano & RPi 4B         & RPi 4B         & Orin Nano      & Xavier NX      & onboard PC \\
AI Accel.          & 67\,TOPS       & 0.5\,TOPS   & 4\,TOPS        & \dsh{}         & 67\,TOPS       & 21\,TOPS       & \dsh{} \\
SLAM-capable       & \cmk{}         & \cmk{}      & \cmk{}         & \cmk{}         & \cmk{}         & \cmk{}         & \cmk{} \\
Open stack         & \cmk{}         & \cmk{}      & \cmk{}         & \cmk{}         & limited        & \cmk{}         & partial \\
Curriculum/docs    & \cmk{}         & \cmk{}      & \cmk{}         & \dsh{}         & \dsh{}         & community      & \dsh{} \\
Maintained product & \cmk{}         & \cmk{}      & community      & \cmk{}         & \cmk{}         & community      & \cmk{} \\
\midrule
Price              & \textbf{\$2{,}700} & \$1{,}098 & $\sim$\$1{,}500$^\dagger$ & \$1{,}850   & \$5{,}690      & \$6{,}000$^*$  & $\sim$\$30{,}000 \\
\bottomrule
\end{tabularx}
{\scriptsize \vspace{0.35em}
\noindent LiDAR is reported as scan frequency (range). ``AI Accel.'' denotes dedicated on-board tera operations per second. RACECAR Neo AI acceleration is provided by a Google Coral Edge TPU (4~TOPS combined with the host).\\
$^*$F1Tenth pricing reflects the RACECAR/J pre-assembled variant~\cite{racecarj}; the official bill of materials~\cite{f1tenthbom} totals approximately \$3{,}800 in parts and requires laser cutting, 3D printing, and multi-vendor assembly.\\
$^\dagger$RACECAR Neo is used at MIT and is not distributed as a product; BWSI publishes an open bill of materials so institutions can source parts and build their own. The figure reflects approximate parts cost.}
\end{table}

\section{Hardware Design}
\label{sec:hardware}

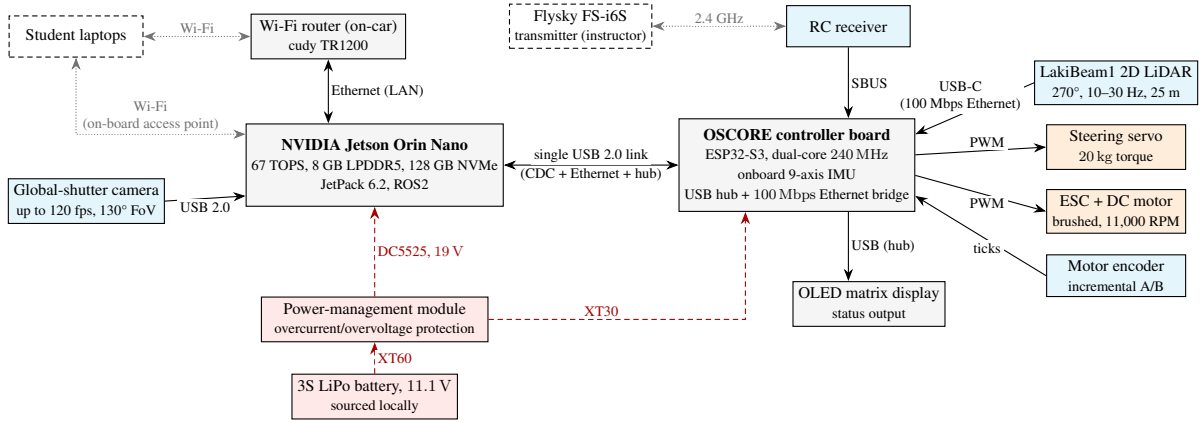
\begin{figure}[t]
\centering
\resizebox{\linewidth}{!}{%
\begin{tikzpicture}[
  font=\footnotesize,
  blk/.style={draw, semithick, align=center, inner sep=3pt, minimum height=8mm, fill=black!4},
  sens/.style={blk, fill=cyan!9},
  act/.style={blk, fill=orange!14},
  pwrn/.style={blk, fill=red!9},
  ext/.style={blk, densely dashed, fill=white},
  data/.style={-{Stealth[length=2.4mm]}, semithick},
  bi/.style={{Stealth[length=2.4mm]}-{Stealth[length=2.4mm]}, semithick},
  rfl/.style={{Stealth[length=2.4mm]}-{Stealth[length=2.4mm]}, semithick, black!55, densely dotted},
  pwr/.style={-{Stealth[length=2.4mm]}, semithick, red!65!black, densely dashed},
  lab/.style={font=\scriptsize, fill=white, inner sep=1.5pt},
]
\node[blk, minimum width=42mm, minimum height=16mm] (jetson) at (0,0)
  {\textbf{NVIDIA Jetson Orin Nano}\\\scriptsize 67 TOPS, 8 GB LPDDR5, 128 GB NVMe\\\scriptsize JetPack 6.2, ROS2};
\node[blk, minimum width=38mm, minimum height=18mm] (oscore) at (8.2,0)
  {\textbf{OSCORE controller board}\\\scriptsize ESP32-S3, dual-core \SI{240}{\mega\hertz}\\\scriptsize onboard 9-axis IMU\\\scriptsize USB hub + \SI{100}{Mbps} Ethernet bridge};
\draw[bi] (jetson.east) -- node[lab, above]{\scriptsize single USB 2.0 link} node[lab, below]{\scriptsize (CDC + Ethernet + hub)} (oscore.west);
\node[sens, minimum width=30mm] (cam) at (-5.6,-0.7) {Global-shutter camera\\\scriptsize up to 120 fps, 130\textdegree{} FoV};
\draw[data] (cam.east) -- node[lab, below]{USB 2.0} ([yshift=-6mm]jetson.west);
\node[ext, minimum width=26mm] (laptop) at (-5.8,2.5) {Student laptops};
\node[blk, minimum width=30mm] (router) at (-0.9,2.5) {Wi-Fi router (on-car)\\\scriptsize cudy TR1200};
\draw[rfl] (laptop.east) -- node[lab, above]{Wi-Fi} (router.west);
\draw[bi] (router.south) -- node[lab, right]{Ethernet (LAN)} (router.south |- jetson.north);
\draw[rfl] (laptop.south) |- node[lab, above, align=center, pos=0.72]{\scriptsize Wi-Fi\\\scriptsize (on-board access point)} ([yshift=6mm]jetson.west);
\node[ext, minimum width=26mm] (tx) at (4.0,2.7) {Flysky FS-i6S\\\scriptsize transmitter (instructor)};
\node[sens, minimum width=24mm] (rx) at (9.2,2.7) {RC receiver};
\draw[rfl] (tx.east) -- node[lab, above]{2.4 GHz} (rx.west);
\draw[data] (rx.south) -- node[lab, right]{SBUS} (rx.south |- oscore.north);
\node[sens, minimum width=27mm] (lidar) at (14.4,1.6) {LakiBeam1 2D LiDAR\\\scriptsize 270\textdegree{}, 10--30 Hz, 25 m};
\node[act, minimum width=27mm] (servo) at (14.4,0.35) {Steering servo\\\scriptsize \SI{20}{\kilogram} torque};
\node[act, minimum width=27mm] (esc) at (14.4,-0.9) {ESC + DC motor\\\scriptsize brushed, 11{,}000 RPM};
\node[sens, minimum width=27mm] (enc) at (14.4,-2.15) {Motor encoder\\\scriptsize incremental A/B};
\draw[data] (lidar.west) -- node[lab, above, align=center, pos=0.62]{\scriptsize USB-C\\\scriptsize (100 Mbps Ethernet)} ([yshift=6mm]oscore.east);
\draw[data] ([yshift=2mm]oscore.east) -- node[lab, above, pos=0.55]{PWM} (servo.west);
\draw[data] ([yshift=-2mm]oscore.east) -- node[lab, below, pos=0.55]{PWM} (esc.west);
\draw[data] (enc.west) -- node[lab, below, pos=0.45]{ticks} ([yshift=-6mm]oscore.east);
\node[blk, minimum width=27mm] (oled) at (9.6,-2.7) {OLED matrix display\\\scriptsize status output};
\draw[data] ([xshift=10mm]oscore.south) -- node[lab, right]{USB (hub)} ([xshift=10mm]oscore.south |- oled.north);
\node[pwrn, minimum width=44mm] (pmm) at (0,-3.0) {Power-management module\\\scriptsize overcurrent/overvoltage protection};
\node[pwrn, minimum width=30mm] (batt) at (0,-4.5) {3S LiPo battery, \SI{11.1}{\volt}\\\scriptsize sourced locally};
\draw[pwr] (batt.north) -- node[lab, right]{XT60} (pmm.south);
\draw[pwr] (pmm.north) -- node[lab, right]{DC5525, \SI{19}{\volt}} (jetson.south);
\draw[pwr] (pmm.east) -| node[lab, above, pos=0.22]{XT30} ([xshift=-10mm]oscore.south);
\end{tikzpicture}%
}
\caption{NeoRacer system architecture. The NVIDIA Jetson Orin Nano hosts perception, planning, and control under ROS2; the OSCORE embedded controller board (ESP32-S3) provides the real-time layer that drives the actuators and reads the sensors requiring microsecond timing. A single USB~2.0 link joins the two computers: the board's onboard hub carries the ESP32's command-and-telemetry channel, the LiDAR scan stream, which arrives over a USB-C cable and is bridged from Ethernet onto the link, and the OLED matrix display, so the camera is the only sensor attached to the Jetson directly. Students reach the vehicle through the on-car Wi-Fi router or the Jetson's own access point. Solid arrows denote data links, dotted arrows wireless links, and dashed red arrows power distribution. The ESC receives raw pack voltage passed through the controller board.}
\label{fig:block_diagram}
\end{figure}

\begin{figure}[h]
\centering
\includegraphics[width=0.95\columnwidth]{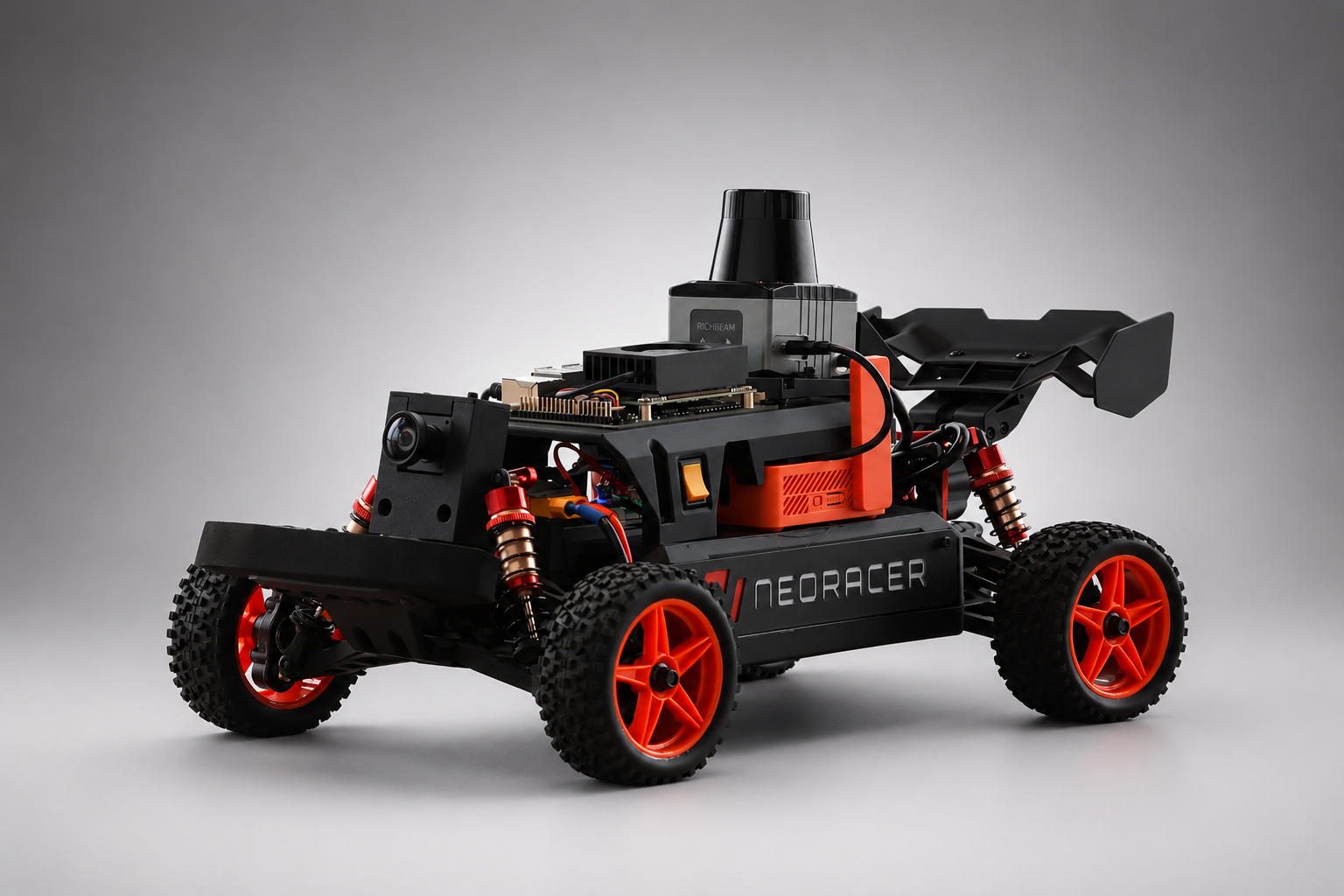}
\caption{The NeoRacer autonomous racing platform. Visible subsystems include the Richbeam LakiBeam1 LiDAR (top centre), the NVIDIA Jetson Orin Nano compute module with active cooling (centre, beneath the LiDAR), the global shutter camera (front left), independent suspension with coilover shocks, the rear spoiler, and the 4WD drivetrain with all-terrain tires.}
\label{fig:platform}
\end{figure}

NeoRacer is a 1:12 scale Ackermann-steered vehicle measuring $380 \times 300 \times 220$~mm and weighing less than \SI{3}{\kilogram}. The platform is designed around five subsystems: compute, perception, actuation, embedded control, and power. The modular architecture allows individual subsystems to be upgraded or replaced independently as technology evolves, without requiring a full platform redesign. Figure~\ref{fig:block_diagram} shows the complete system architecture, including the two computing systems, their interconnection, and the connections to every sensor and actuator; Figure~\ref{fig:platform} shows the assembled platform.

\subsection{Compute Module}

The primary compute unit is an NVIDIA Jetson Orin Nano mounted on a Seeed Studio J401 carrier board. The Orin Nano provides 67~TOPS of AI inference performance with an Ampere-architecture GPU, \SI{8}{\giga\byte} LPDDR5 RAM, and hardware acceleration for CUDA, cuDNN, and TensorRT. Storage is provided by a \SI{128}{\giga\byte} NVMe SSD.

The module ships pre-flashed with JetPack 6.2, an Ubuntu-based Linux distribution (L4T) that includes CUDA 12.x, cuDNN, TensorRT, and a complete ROS2 development environment. Four USB~3.2 Type-A ports, Gigabit Ethernet, DisplayPort~1.2, and a 40-pin GPIO header provide connectivity to all peripherals.

This compute capability is a defining differentiator. At 67~TOPS, the Orin Nano can simultaneously run LiDAR-based SLAM (e.g., Cartographer), camera-based object detection (e.g., YOLOv8), and a real-time control loop at \SI{30}{\hertz}. This workload exceeds the capacity of the Jetson Xavier NX (21~TOPS) used in F1Tenth and is infeasible on the Raspberry Pi 4 ($<$1~TOPS) used in lower-cost platforms.

\subsection{Perception Suite}

\begin{figure}[h]
\centering
\includegraphics[width=0.75\columnwidth]{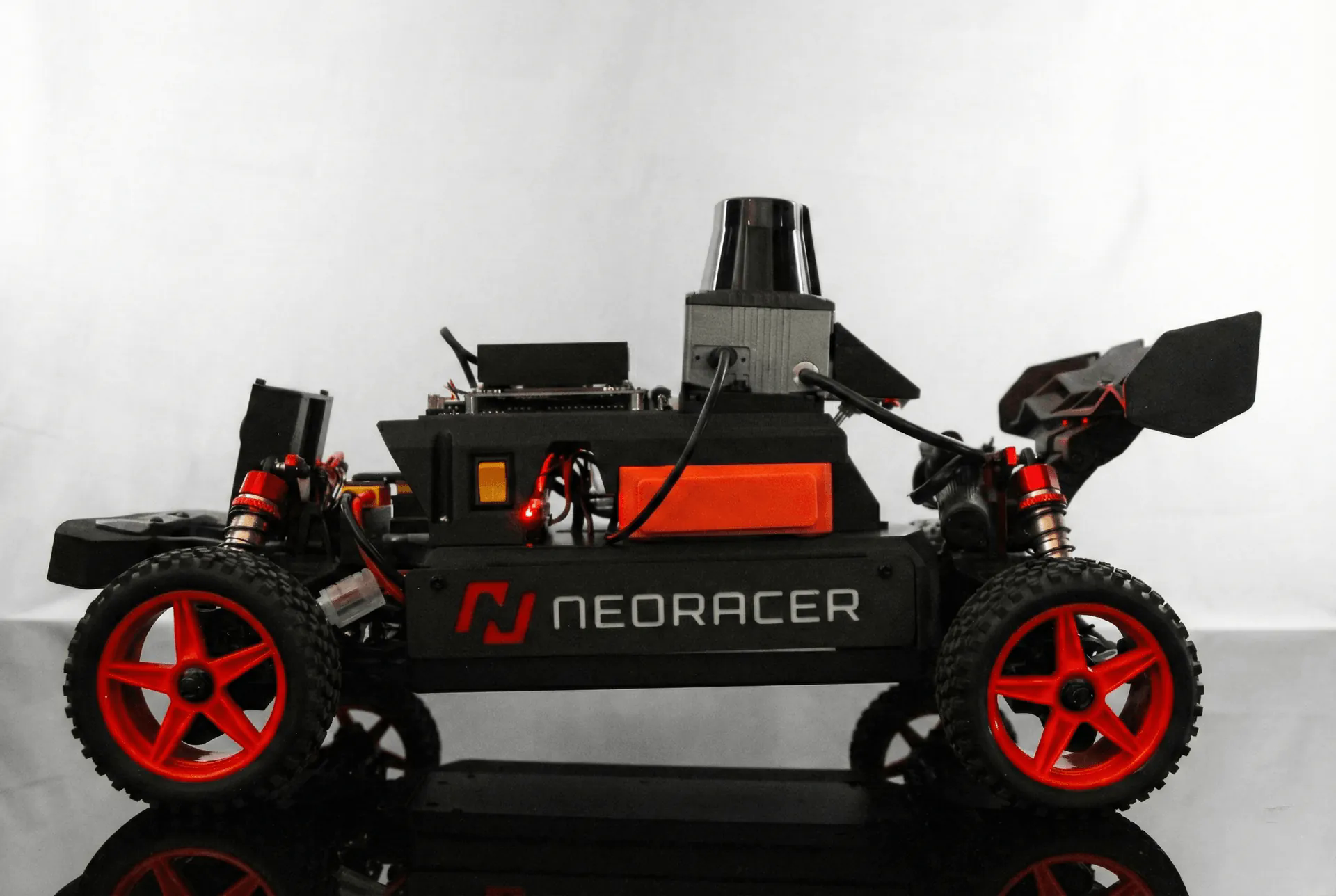}
\caption{Side profile showing the LiDAR (cylindrical unit, top), camera module (left), independent coilover suspension, and NeoRacer branding on the 6061 aluminum chassis.}
\label{fig:profile}
\end{figure}

\paragraph{2D LiDAR.}
The Richbeam LakiBeam1 provides 270\textdegree{} horizontal coverage with a configurable scan rate of 10--30~Hz and angular resolution of 0.08\textdegree--0.24\textdegree{}. Range extends to \SI{25}{\meter} at 70\% reflectivity (\SI{15}{\meter} at 10\%), with \SI{\pm 2}{\centi\meter} accuracy. The unit connects to the embedded controller board over a USB-C cable carrying its 100~Mbps Ethernet interface; the board bridges the scan stream onto the single USB link to the Jetson (\S\ref{sec:embedded}). The sensor is rated IP67 for dust and water resistance. At $55 \times 55 \times 51$~mm and \SI{85}{\gram}, it integrates cleanly into the vehicle's form factor; Figure~\ref{fig:profile} shows the sensor placement in side profile.

\paragraph{Global Shutter Camera.}
A 2.3~MP global shutter camera captures $1920 \times 1200$ frames at 90~fps (full resolution) or $1280 \times 720$ at 120~fps in racing mode. The 130\textdegree{} field of view with a 2.7~mm low-distortion lens provides wide forward coverage. Global shutter eliminates rolling shutter artifacts during high-speed motion, which is critical for reliable lane detection and AprilTag tracking at speeds up to \SI{6}{\meter\per\second}. The camera was upgraded from 30~fps to 120~fps based on findings from the MIT IAP pilot (\S\ref{sec:evaluation}), where high-speed racing revealed that 30~fps introduced unacceptable latency in the perception pipeline.

\paragraph{Inertial Measurement Unit.}
A 9-axis IMU is integrated directly on the embedded controller board (\S\ref{sec:embedded}): a QMI8658A 6-axis accelerometer/gyroscope paired with a QMC6309 3-axis magnetometer on a shared I\textsuperscript{2}C bus. The ESP32-S3 reads both parts and provides AHRS (Attitude and Heading Reference System) fusion for orientation estimation and dead reckoning.

\paragraph{Motor Encoder.}
Odometry feedback comes from an incremental encoder integrated into the custom brushed drive motor rather than from per-wheel sensors. The encoder provides quadrature A/B channels at a resolution of up to 4096 counts per revolution and is rated to 20{,}000~rpm, well above the motor's operating range. Reading the two phase-shifted channels recovers direction as well as speed, and the embedded controller integrates the pulse counts, through the fixed drivetrain ratio, into a velocity estimate that is published as standard ROS2 odometry. This dead-reckoning signal bridges the gap between LiDAR scans and closes the loop for velocity control.

\subsection{Actuation}

The drivetrain uses a single custom brushed DC motor (11,000~RPM no-load) with an integrated shaft encoder, driving all four wheels through a shaft-drive 4WD system. An electronic speed controller (ESC) provides precise PWM-based speed and torque regulation. Steering is provided by a \SI{20}{\kilogram}-torque waterproof servo in an Ackermann geometry configuration with a \SI{280}{\milli\meter} wheelbase.

\begin{figure}[h]
\centering
\includegraphics[width=0.48\columnwidth]{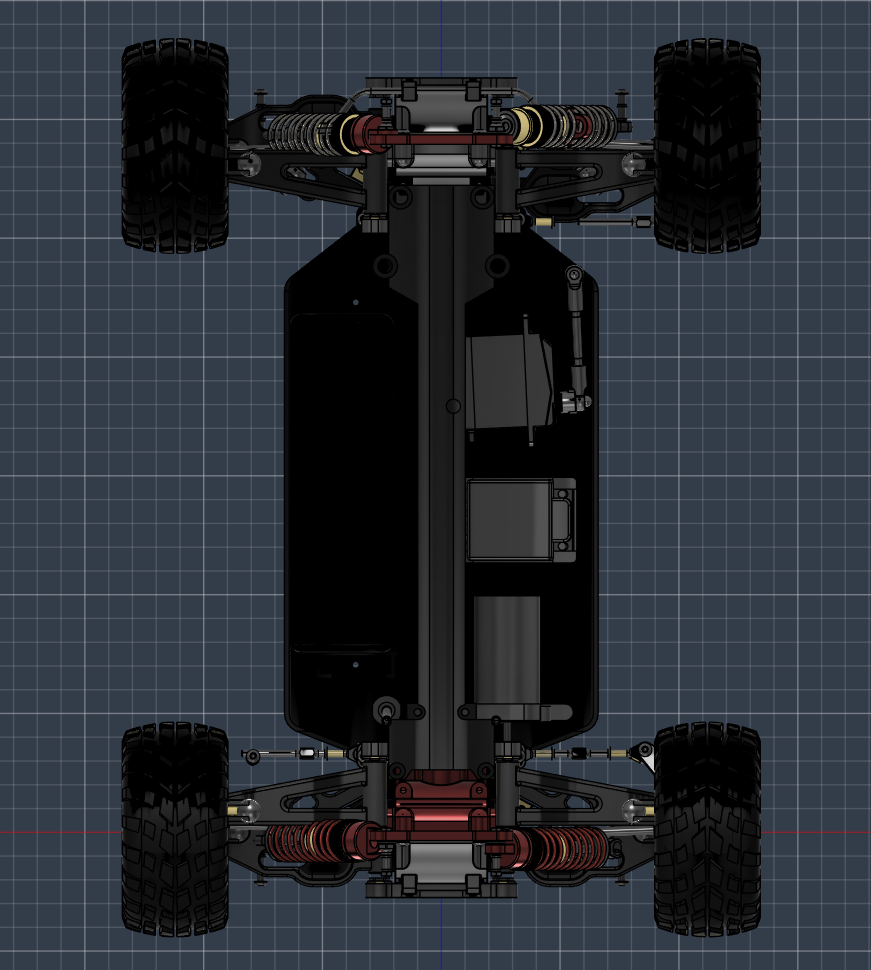}
\caption{Top-down CAD view of the 6061 aluminum alloy chassis, showing the independent suspension geometry at all four corners and the shaft-drive 4WD drivetrain.}
\label{fig:topdown}
\end{figure}

Independent aluminum alloy spring suspension on all four corners provides consistent tire contact on uneven surfaces. All-terrain tires ($<$\SI{80}{\milli\meter} diameter) and anti-collision foam bumpers enhance durability in classroom and competition environments. Figure~\ref{fig:topdown} shows the chassis layout, suspension geometry, and drivetrain from above.

\subsection{Embedded Controller}
\label{sec:embedded}

The one piece of custom electronics in the vehicle is OSCORE, a robot control board built around an ESP32-S3-WROOM-1U module (dual-core Xtensa LX7 at \SI{240}{\mega\hertz}). While the Jetson runs the autonomy stack, OSCORE is the real-time layer underneath it: it generates the PWM signals that drive the ESC and the steering servo, reads the motor encoder, samples the onboard 9-axis IMU, and accepts the SBUS stream from the RC receiver. Placing these deadline-critical loops on a dedicated microcontroller isolates them from scheduling jitter on the Linux side, so a heavily loaded perception pipeline can never delay a steering update or a stop command.

The board communicates with the Jetson over a single USB~2.0 link. An onboard hub (CH339F) sits behind that link and aggregates three downstream paths: the ESP32-S3's USB-CDC command and telemetry channel, a \SI{100}{Mbps} Ethernet bridge that carries the LiDAR's UDP scan stream, and USB ports serving the OLED matrix status display and free expansion connectors. A CAN bus output is available for additional peripherals. A two-stage power section converts the 9--26~V input into regulated \SI{5}{\volt} (\SI{5}{\ampere}) and \SI{3.3}{\volt} rails, with the raw pack voltage passed straight through to the ESC and reverse-polarity protection on the input. Status indication is provided by a programmable WS2812 RGB LED and an onboard buzzer. The board also implements a safety watchdog: if the Jetson stops sending commands, the ESP32 triggers a controlled stop.

The full electrical documentation for OSCORE (the hardware manual, complete schematic, and reference designator map) is published under CERN-OHL-S~v2 (\S\ref{sec:licensing}). The board was redesigned following the MIT pilot to improve reliability under sustained high-current loads during extended racing sessions (\S\ref{sec:evaluation}).

\subsection{Power System}

NeoRacer runs a single hierarchical power path from one 3S (11.1~V) LiPo battery, with a recommended specification of 5200~mAh capacity at 50C discharge rate (minimum 25C), connected via XT60. The battery is not included with the kit due to international shipping regulations for lithium-ion cells; users source batteries locally, which also allows selection of capacity appropriate to their use case. The battery feeds a power-management module that distributes every downstream rail (Fig.~\ref{fig:block_diagram}): a regulated \SI{19}{\volt} DC5525 line to the Jetson, an XT30 line to the embedded controller, and a \SI{5}{\volt} USB bus to the sensors. The module provides overcurrent and overvoltage protection so that a fault on one rail does not take down the rest of the vehicle.

\subsection{Networking and Remote Access}

Students have two ways to reach the car, and both create a network wherever the car is deployed rather than depending on institutional Wi-Fi, a practical concern in the gyms, hallways, and competition venues where scaled racing actually happens. The first is a compact Wi-Fi~6 travel router (cudy TR1200) carried on the vehicle and wired to the Jetson's Gigabit Ethernet port. The second is the Jetson's own wireless interface running in access-point mode, in which the car broadcasts its own network and laptops join it directly with no additional hardware. In either case students reach the Jetson at a fixed address over SSH or a remote desktop session, and ROS2 DDS discovery works across the same link, so live sensor topics can be visualized off-board without configuration.

\subsection{Safety Features}

Safety in a classroom environment with high-speed (\SI{6}{\meter\per\second}) vehicles is paramount, and NeoRacer layers protections at the mechanical, radio, software, and firmware levels so that no single failure leaves the vehicle uncontrolled.

At the mechanical level, anti-collision foam bumpers and the spoiler absorb impact energy in the inevitable early-lab wall contacts, and pinion gear protection prevents drivetrain damage from sudden lockups. At the radio level, every kit ships with a 2.4~GHz Flysky transmitter carrying two dedicated toggle switches whose state the embedded controller reads from the SBUS frame on every control cycle. One switch selects the control source: in its manual position the sticks have full control and autonomous commands are ignored, and only in its autonomy position does the ROS2 stack drive the vehicle. The other switch caps manual driving at 15\% speed for supervised operation. Because this arbitration runs in firmware rather than on the Linux host, an instructor holding the transmitter can reclaim control of any student vehicle instantly, from across a room, without touching a keyboard; if the radio link is lost entirely, the firmware detects the receiver's failsafe flag, brings the throttle to neutral, and returns the steering to center. At the software level, a LiDAR-based safety-stop module provides emergency braking when an obstacle enters the vehicle's path. Finally, at the firmware level, the ESP32 watchdog timer described in \S\ref{sec:embedded} halts the vehicle if the command stream from the Jetson stops for any reason: a crashed node, a dropped process, or a wedged operating system.

\section{Release Identity and Manufacturing Traceability}
\label{sec:traceability}

For an open hardware platform to function as shared infrastructure rather than a single-lab artefact, each shipped unit must be a controlled physical object whose hardware revision, software image, and manufacturing context can be identified after the fact. A user that publishes a result, files a bug report, or requests a replacement part should be able to identify the exact configuration on which the work was performed. NeoRacer is co-developed by the Neobotics Foundation and Seeed Studio and manufactured by Seeed Studio; consolidating production with a single manufacturer is what makes the traceability scheme described here enforceable across batches. We address this by coupling each NeoRacer unit to three independent identifiers and by structuring production around a fixed release loop.

\subsection{Three-Identifier Release Scheme}

Each shipped NeoRacer carries three identifiers that together constitute its release identity.

\begin{itemize}
  \item \textbf{Hardware revision:} \texttt{NR-HW-v\textit{major}.\textit{feature}.\textit{patch}} identifies the physical platform generation. A major bump signals a breaking change to chassis, compute, or sensor interfaces; feature and patch bumps cover non-breaking additions and tolerance corrections.
  \item \textbf{Software release:} \texttt{nrlib-v\textit{major}.\textit{minor}.\textit{patch}} identifies the validated software stack associated with the unit, including drivers, ROS2 packages, simulation assets, and documentation.
  \item \textbf{Manufacturing run:} \texttt{MFG-\textit{YYYY}-W\textit{ww}-\textit{vendor}-\textit{SHA}} identifies the year, ISO week, fabrication vendor, and a short content hash over the bill of materials and firmware image used for the batch.
\end{itemize}

A reproducibility statement in a paper, lab notebook, or course handout can therefore identify the exact hardware, software, and manufacturing context in a single line (for example, \texttt{NR-HW-v2.1 / nrlib-v3.4 / MFG-2026-W18-seeed-a1f3e2}). This makes hardware drift, software drift, and batch-level part substitutions visible across institutions and over time. Each software release ships with a compatibility matrix that classifies hardware/software pairings as validated, compatible with known limitations, or unsupported, so users can determine in advance whether a given controller, driver, or simulation asset is expected to work against their unit.

\subsection{Release Loop}

Production proceeds in a fixed four-stage loop that turns each batch into a controlled release rather than a one-time build.

\begin{enumerate}
  \item \textbf{Release freeze.} The bill of materials, assembly procedure, acceptance-test checklist, software image, replacement-part plan, and accompanying documentation are frozen and tagged. This produces a known hardware release candidate before manufacturing begins, so any later substitution or process change is explicit rather than implicit.
  \item \textbf{Batch build.} Seeed Studio fabricates and assembles units against the frozen specification. Part substitutions, build-condition deviations, and assembly QA failures are recorded against the batch identifier. Early batches in particular surface long-lead parts, sensor and compute-module substitutions, and post-shipment failure modes.
  \item \textbf{Validation.} Each batch is validated against the corresponding software image, documentation, and acceptance tests before release approval. The output is an approved hardware/software pairing, an updated compatibility matrix entry, and a per-batch replacement-part record for institutional maintenance.
  \item \textbf{Community feedback.} Issues raised by pilot users and course deployments feed the next release freeze. This closes the loop between manufacturing, field use, and platform improvement, and prevents informal fixes from accumulating outside the version-controlled release path.
\end{enumerate}

\subsection{Practical Implications}

The release identity scheme has three practical consequences. First, results obtained on NeoRacer can be reproduced across institutions because the configuration is identifiable in citation form. Second, replacement parts can be ordered against a known hardware revision and validated software release without ad-hoc compatibility checks. Third, support traffic is bounded: when a user reports an issue, the release identity narrows the search space to a known set of hardware and software, rather than requiring case-by-case reconstruction of the originating environment. The scheme is intentionally lightweight; it adds three short strings to the documentation that ships with each unit and does not require dedicated tooling beyond a version-controlled manufacturing record.

\section{Software Architecture}
\label{sec:software}

\subsection{Operating System and Framework}

NeoRacer ships with JetPack 6.2 pre-installed, providing Ubuntu Linux with ROS2 (Robot Operating System 2) as the primary middleware. ROS2's DDS-based publish-subscribe architecture enables modular development: users write individual nodes for perception, planning, and control that communicate via typed topics, fostering software engineering best practices.

The primary development language is Python 3.10+, chosen for its accessibility to students with introductory programming experience. C++ is supported for performance-critical nodes.

\subsection{Pre-Configured Drivers}

All hardware drivers are pre-installed, tested, and exposed as ROS2 nodes against the standard topic and message conventions summarised in Table~\ref{tab:drivers}. A single \texttt{controller} node owns the USB link to the OSCORE board (\S\ref{sec:embedded}): it publishes the IMU, odometry, and RC transmitter state, and writes actuation commands back to the ESP32. This pre-configuration eliminates the multi-week driver setup phase that is common with other platforms, allowing users to begin algorithm development immediately upon unboxing.

\begin{table}[h!]
\centering
\caption{Pre-configured ROS2 drivers shipped with NeoRacer.}
\label{tab:drivers}
\renewcommand{\arraystretch}{1.15}
\footnotesize
\begin{tabular}{@{}lllp{4.3cm}@{}}
\toprule
\textbf{Node} & \textbf{Direction} & \textbf{Topic} & \textbf{Message type} \\
\midrule
\texttt{richbeam\_driver} & publisher   & \texttt{/scan}      & \path{sensor_msgs/LaserScan} \\
\texttt{camera\_driver}   & publisher   & \texttt{/camera}    & \path{sensor_msgs/Image} \\
\texttt{controller}       & publisher   & \texttt{/imu}       & \path{sensor_msgs/Imu} \\
\texttt{controller}       & publisher   & \texttt{/odom}      & \path{nav_msgs/Odometry} \\
\texttt{controller}       & publisher   & \texttt{/joy}       & \path{sensor_msgs/Joy} \\
\texttt{controller}       & subscriber  & \texttt{/drive}     & \path{ackermann_msgs/AckermannDriveStamped} \\
\bottomrule
\end{tabular}
\end{table}

\subsection{SLAM and Navigation Compatibility}

The sensor suite and compute capability support standard SLAM and navigation stacks including Google Cartographer, SLAM Toolbox, and Nav2. TensorRT-accelerated inference enables real-time object detection (YOLOv8) alongside SLAM, a workload combination that is infeasible on platforms with less than 20~TOPS of compute.

\subsection{Stable Programming Interface}
\label{sec:stable_api}

User-facing code targets \texttt{racecar-neo-library}, a Python interface released under GPLv3 and originally derived from the \texttt{racecar\_core} library used in the MIT RACECAR program~\cite{mitracecar2017} and its Beaver Works Summer Institute (BWSI) instructional materials. The library exposes high-level access to sensors, actuators, and timing under a stable API. When a sensor model or compute module is substituted in a later production run, the corresponding driver is updated inside the library while student code, curriculum modules, and simulator examples continue to target the same calls. This separation enables long-term curriculum stability and cross-cohort reproducibility: a course or experiment written against \texttt{nrlib-v3.x} does not break merely because the hardware generation has advanced.

\subsection{Simulation Environment}
\label{sec:simulation}

\begin{figure}[h!]
\centering
\includegraphics[width=0.95\linewidth]{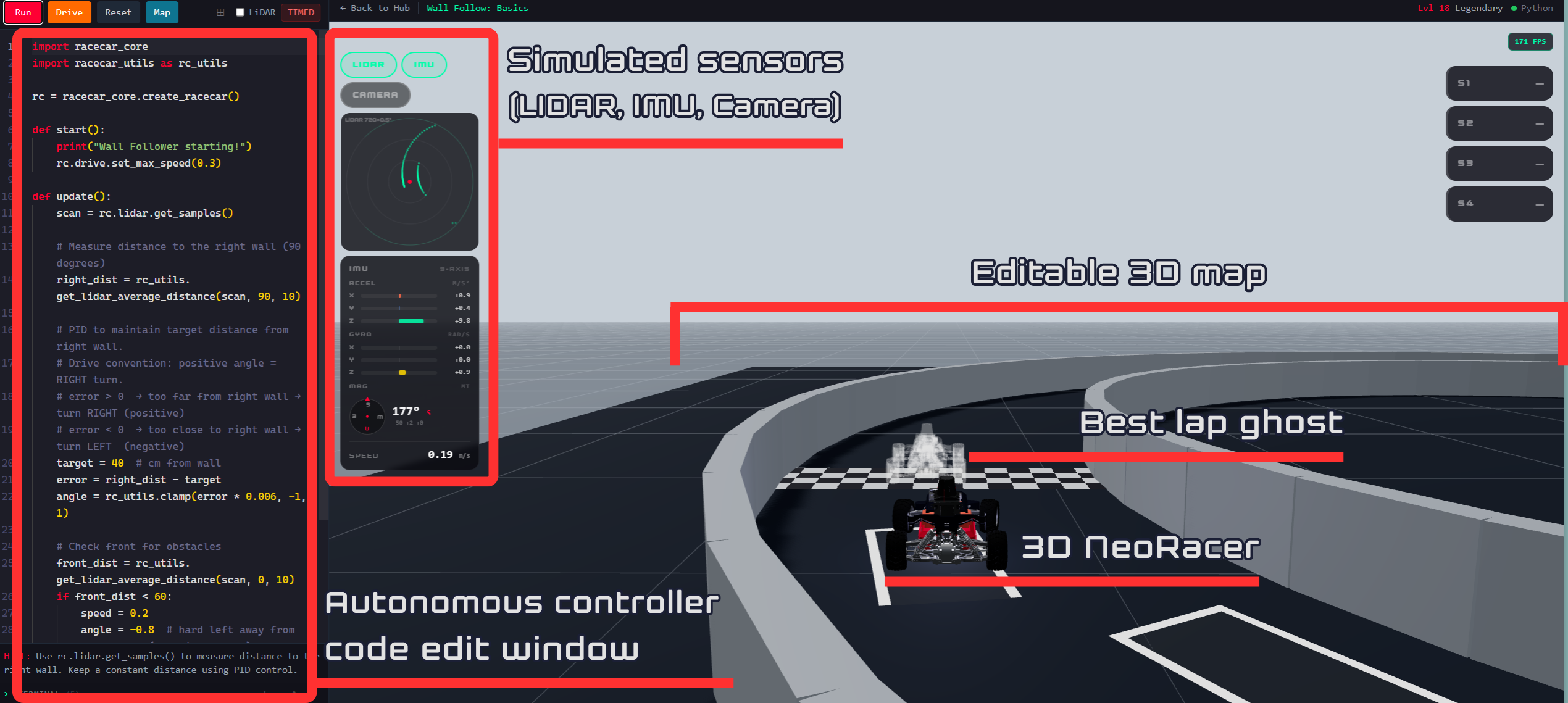}
\caption{The Neobotics Playground in-browser simulator. The interface combines a controller-code editor (left), live sensor readouts (centre), and a three-dimensional simulated NeoRacer on an editable track with a best-lap ghost (right). Code written against this interface runs unchanged on the physical car.}
\label{fig:playground}
\end{figure}

\begin{figure}[h!]
\centering
\includegraphics[width=0.6\linewidth]{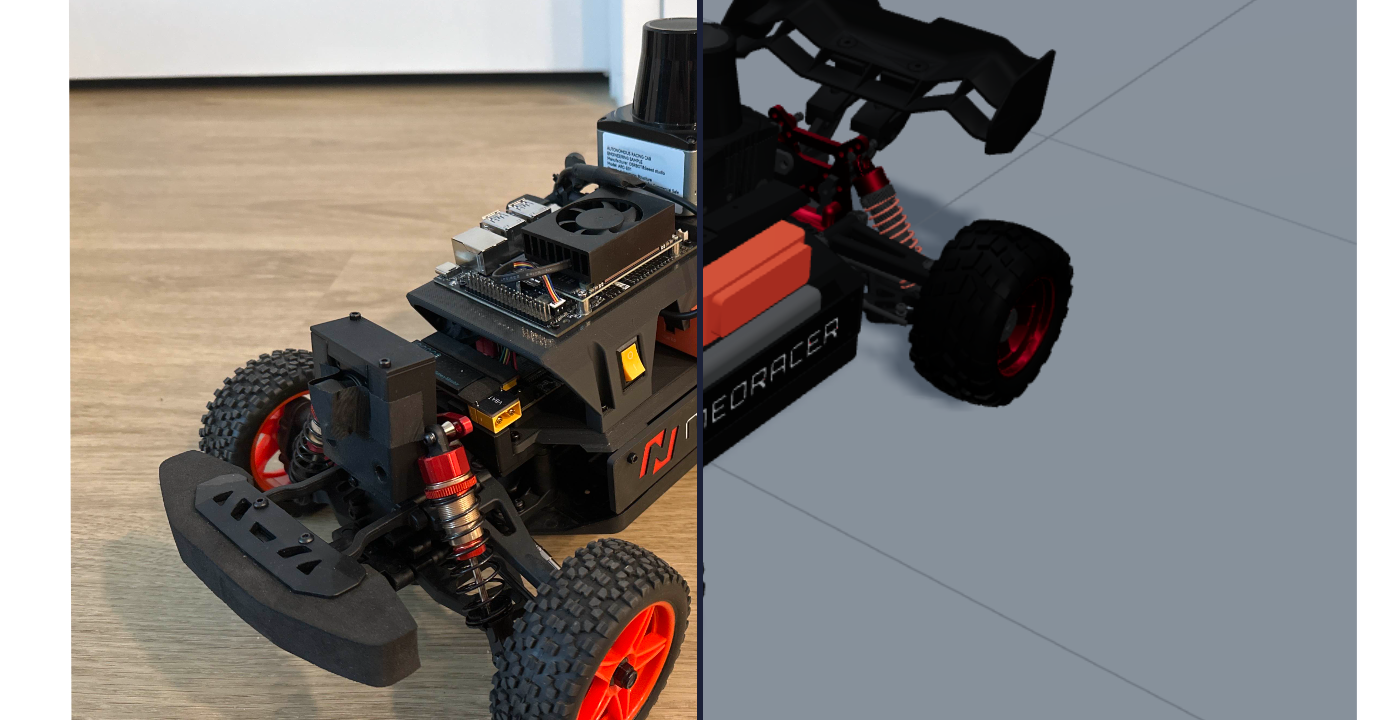}
\caption{Matched physical and simulated views of the NeoRacer. The same vehicle model, sensor placements, and dimensions are used in both the physical platform (left) and the Playground three-dimensional model (right). This correspondence is the basis for sim-to-real algorithm transfer.}
\label{fig:real_sim}
\end{figure}

The primary simulation environment is the Neobotics Playground, an in-browser three-dimensional simulator built on a physically calibrated model of the NeoRacer (Fig.~\ref{fig:real_sim}). The Playground exposes the same \texttt{racecar-neo-library} API used on the physical car, edits and runs autonomous controllers in Python, simulates LiDAR, IMU, and camera streams, and supports user-editable tracks and recorded best-lap ghosts (Fig.~\ref{fig:playground}). Because the simulator and the physical platform target the same library, code developed in the browser can be transferred to the car without modification; this is the same property that virtual qualifying rounds in the competition framework (\S\ref{sec:competition}) rely on.

For research users who require alternative simulation backends, the project maintains URDF-style robot descriptions, controller interfaces, and example shims that connect the same hardware baseline to MuJoCo, Isaac Lab, and Gazebo. This keeps the Playground as the maintained primary path for beginners and instructors while preserving the ability to run the same physical NeoRacer against specialised simulation backends for research workloads. Differences between simulated and physical behaviour can be measured against shared logs and reported as a sim-to-real gap, which is itself useful as a debugging signal, a teaching example, and a research benchmark.

\subsection{Standardized Benchmarking}

Because every NeoRacer ships with identical hardware, sensor configurations, and driver interfaces, the platform provides a controlled environment for benchmarking autonomous racing algorithms. Researchers at different institutions can compare SLAM implementations, path planners, or control strategies on hardware-equivalent platforms, eliminating the confounding variable of heterogeneous sensor suites and compute capabilities that plagues cross-institutional comparison in the current landscape.

\section{Research Enablement}
\label{sec:research}

\begin{objbox}
\noindent\textbf{Objective.} The hardware specification (\S\ref{sec:hardware}), the traceable release identity (\S\ref{sec:traceability}), and the stable programming interface (\S\ref{sec:stable_api}) together make NeoRacer a controlled substrate against which existing autonomous-systems research can be re-evaluated on shared hardware, and against which new methods can be benchmarked across institutions. This section identifies four research directions for which the platform is a particularly well-matched testbed.
\end{objbox}

\Paragraph{Learning-Enabled Control on Embedded Hardware.}
Reinforcement-learning-based controllers for vehicle dynamics, racing-line tracking, and obstacle avoidance routinely succeed in simulation and then degrade on hardware because of actuation limits, model mismatch, sensor noise, and timing variability. Sim-to-real frameworks such as RE+AL~\cite{how_to_train_your_quadrotor} and CAPS~\cite{caps} have addressed this gap for quadrotor platforms. Subsequent work on Fulfillment Priority Logic~\cite{fpl}, asymmetric actor-critic architectures~\cite{asymetric_actor_critic}, and live anchored adaptation~\cite{swannflight} has extended the methodology to multi-objective deployment with bounded forgetting. The NeoRacer compute budget, deterministic embedded controller path, and matched sim-to-physical correspondence (\S\ref{sec:simulation}) provide a directly comparable testbed for these methods on a ground vehicle. Because every unit ships with identical sensing and compute, controllers released for one NeoRacer can be re-evaluated on another at a different institution without confounding hardware differences.

\Paragraph{Stability-Certified Neural Control.}
The deployment of learned controllers in safety-critical settings has motivated a body of work on verifiable certificates, including recent Lyapunov-stable neural controllers accelerated through Fulfillment Priority Logic~\cite{abdelgawad2026lyapunov}. Validating these methods requires a platform on which controller assumptions, closed-loop physical behaviour, and certificate operating regions can all be observed concurrently. NeoRacer supports this validation loop: controllers can be developed and certified in simulation against the same robot description used on the physical car, deployed against a known hardware revision, and evaluated against logs that can be shared in a reproducible form using the release identity scheme.

\Paragraph{Real-Time Resource Management on Heterogeneous SoCs.}
Modern autonomy workloads place perception, SLAM, learning inference, planning, and control on the same multi-core, accelerator-equipped SoC. The resulting contention for memory bandwidth, cache, and I/O is increasingly the dominant determinant of closed-loop timing. The Omnivisor real-time partitioning hypervisor extension~\cite{Ottaviano:ECRTS24}, the E-WarP profiling and bandwidth-management framework~\cite{Sohal:RTSS20}, and microsecond-scale memory regulation through MemPol~\cite{Zuepke:RTAS23} have been developed in this context. Most existing evaluations of these techniques target synthetic workloads or bench setups. NeoRacer provides a physically grounded, closed-loop platform on which the deadline-critical effects of resource contention become visible at the level of lap time, tracking error, and stability, rather than only microbenchmark counters. The progress-aware management approach formalised in CAPA~\cite{Chen:TC26} is a natural target for deployment on the platform.

\Paragraph{Hardware/Software Co-Design.}
The modular compute carrier on NeoRacer admits alternative compute paths beyond the default Jetson Orin Nano, including FPGA+CPU configurations on which monitoring, scheduling, and acceleration primitives can be placed directly in programmable logic~\cite{Mancuso:CPSIOT23}. This makes the platform suitable as a target for hardware/software co-design studies in which an autonomy stack is repartitioned between general-purpose cores and reconfigurable fabric, evaluated under racing-grade timing constraints, and compared against the standard Orin Nano baseline on the same chassis and sensing suite.

The directions above are not the focus of this paper. They are included to clarify that the platform documented here is intended to function not only as a teaching kit but as a controlled substrate against which the autonomous-systems research community can compare results without first reconciling differences in robots.

\section{Reference Curriculum}
\label{sec:curriculum}

\begin{objbox}
\noindent\textbf{Objective.} A reference curriculum is in active development around the platform. The objective is not to ship a single fixed course, but to demonstrate that a single physical baseline can be used coherently across a wide range of experience levels, from high school students with no prior robotics exposure through graduate students working on autonomy research. The platform itself addresses product stability through the hardware design (\S\ref{sec:hardware}) and the release identity scheme (\S\ref{sec:traceability}), and addresses transparency through a non-opaque, fully inspectable software stack (\S\ref{sec:software}); the curriculum addresses the residual challenge of pedagogical-complexity scaling across this user range.
\end{objbox}

NeoRacer does not currently ship with a bundled curriculum. A reference curriculum is being developed alongside the platform, and the design target is for course content, lessons, assignments, and gated challenges to live inside the Neobotics Playground (\S\ref{sec:simulation}) rather than as a separate document set. This places the same simulator that students use to write and test controllers in the same place as the learning material, the progress tracking, and the assignment submission path, so that simulator runs, hardware deployments, and curriculum progress remain tied to the same release identity (\S\ref{sec:traceability}).

\Paragraph{Course Structure.}
The reference course design targets a 12-week, 4-credit university course comprising two 75-minute lectures and one 50-minute lab per week. The target class size is 48 students organized into 12 teams of 4. Prerequisites are limited to introductory Python programming (CS1/CS2 equivalent). A condensed 2-week intensive variant was used for the MIT Independent Activities Period (IAP) pilot (\S\ref{sec:evaluation}).

\Paragraph{Assessment Model.}
The reference grading scheme comprises weekly labs (40\%, pass/fail gate challenges), a midterm demonstration (20\%, simulation-based autonomous navigation), a final race project (30\%, full autonomous racing on hardware), and class participation (10\%). An experience point (XP) layer provides bonus credit for fastest laps, code quality, and peer assistance, encouraging both competition and collaboration. Because gate challenges and assessments are designed to run inside the Playground, the same scaffolding can support shorter intensives (such as the MIT IAP variant) without redesigning the assessment instruments.

\section{Cost Analysis}
\label{sec:cost}

\subsection{Platform Cost}

NeoRacer is priced at \$2,700 per unit at retail, with early-bird pre-order pricing at \$2,500 and institutional volume pricing that goes down to \$2,200 per unit for commitments of 15 or more units. At institutional pricing, a \$26,400 budget equips 12 teams with NeoRacers pre-assembled and ready to use. The same budget procures only four pre-assembled F1Tenth units at \$6,000 each, or 24 DuckieBots with 0.5~TOPS, which cannot run modern perception algorithms at racing speeds. The headline per-unit numbers and accompanying feature support are summarised in the platform comparison of \S\ref{sec:related} (Table~\ref{tab:comparison}) and are not duplicated here.

\subsection{Sponsor-a-Kit Program}

To address equity in access, Neobotics operates a Sponsor-a-Kit program through which corporate and individual sponsors fund kits for under-resourced schools at \$3,200 per kit (covering hardware, shipping, setup, and onboarding). As a 501(c)(3) nonprofit, all contributions are tax-deductible. This model is inspired by the FIRST Robotics team sponsorship ecosystem and is designed to scale platform access beyond institutions with existing robotics budgets.

\section{Pilot Deployment}
\label{sec:evaluation}

A 2-week intensive pilot was conducted during MIT's January--February Independent Activities Period (IAP) session with 15 undergraduate students. The compressed timeline required students to progress from unboxing to autonomous racing in 10 working days, providing a stress test of both the hardware's reliability and the platform's accessibility to students with no prior robotics experience.

The MIT pilot yielded two critical hardware revisions:

\begin{enumerate}
  \item \textbf{Power board redesign:} Sustained high-current loads during extended multi-car racing sessions exposed reliability issues in the original power distribution board. The board was redesigned with improved current handling and thermal management.
  \item \textbf{Camera upgrade (30~fps $\rightarrow$ 120~fps):} At racing speeds approaching \SI{6}{\meter\per\second}, the original 30~fps camera introduced unacceptable latency in the perception pipeline. Frame-to-frame displacement at full speed exceeded the camera's ability to provide temporally coherent input for lane detection and obstacle avoidance. The camera was upgraded to a 120~fps global shutter module, reducing per-frame latency from \SI{33}{\milli\second} to \SI{8.3}{\milli\second}.
\end{enumerate}

These revisions demonstrate the value of early pilot deployments in a demanding environment: issues that would not surface in bench testing became immediately apparent under competitive racing conditions.


Table~\ref{tab:revisions} summarizes the hardware revisions made as a result of pilot feedback.

\begin{table}[h!]
\centering
\caption{Hardware revisions informed by pilot deployments.}
\label{tab:revisions}
\begin{tabular}{@{}llll@{}}
\toprule
\textbf{Component} & \textbf{Issue} & \textbf{Resolution} & \textbf{Source} \\
\midrule
Power board   & Thermal failure under load     & Redesigned with higher current rating & MIT IAP \\
Camera        & 30~fps insufficient at speed   & Upgraded to 120~fps global shutter    & MIT IAP \\
\bottomrule
\end{tabular}
\end{table}

\section{Toward a Global Competition Framework}
\label{sec:competition}

The standardized, spec-locked nature of NeoRacer enables a competition format where performance differences are attributable to software and algorithmic innovation rather than hardware advantages. We outline a planned competition framework inspired by FIRST Robotics and the F1Tenth Grand Prix series.

\Paragraph{Virtual Qualifying.}
Teams submit autonomous driving code to a standardized simulation environment, and top performers advance to in-person events. Because the simulator exposes the same programming interface as the physical car (\S\ref{sec:simulation}), qualifying code is directly transferable to hardware rather than a separate artifact. This stage eliminates travel barriers and enables global participation at zero marginal cost: a team without access to a physical NeoRacer can still compete, and a school evaluating the platform can enter the ecosystem before purchasing hardware.

\Paragraph{Regional Events.}
University robotics clubs host regional competitions using a published Competition Host Kit containing track specifications, the timing system, scoring rules, and branding assets. In this affiliate model, Neobotics provides the platform and rules while partners provide venues and volunteers, which enables scaling without centralized operational cost, the same structural property that allowed FIRST to grow from a single event to a global league.

\Paragraph{Championship Finals.}
Top qualifiers from regional events and virtual qualifying compete at an annual championship hosted at a major research university, giving the season a visible terminal event and giving participating institutions a recurring reason to sustain their teams year over year.

\Paragraph{Visual Customization.}
Teams may visually customize their vehicles (livery, bodywork) but may not modify hardware specifications. This preserves the benchmarking property that motivates the spec-locked design (every car on the grid is mechanically identical) while still giving teams the identity-building outlet that drives engagement in established robotics leagues.

This structure is designed to grow incrementally. The first flagship event is planned at the 2026 IEEE Vehicular Technology Conference (VTC 2026) in Boston, co-organised with the RoboRacer Foundation~\cite{f1tenth2020} as a multi-school invitational; that event does not follow the framework proposed here. The Neobotics competition format itself is still in the ideation phase, and its rules, scoring, and event structure remain to be decided. Subsequent years extend to 3--5 university-hosted regionals in 2027 and to international expansion in 2028--2029.

\section{Broader Impacts and Sustainability}
\label{sec:broader}

\Paragraph{Workforce and Access.}
NeoRacer reduces the financial and technical entry barriers to working with research-grade autonomous-systems hardware. The compute, sensing, and software stack target the heterogeneous architectures now standard in the autonomous-vehicle industry, including the NVIDIA Orin family. Students working on NeoRacer in classroom or club settings are therefore exposed to the same compute substrate they encounter in subsequent industry or research roles, rather than to platform-specific abstractions that do not generalise. Combined with the open licensing scheme (\S\ref{sec:licensing}) and the Sponsor-a-Kit program (\S\ref{sec:cost}), this lowers the cost of entry both for individual institutions and for under-resourced schools.

\Paragraph{Pathway from Simulation to Physical Competition.}
The outreach strategy is organised as a continuous pathway rather than as isolated workshops. Participants can begin in the in-browser Playground simulator (\S\ref{sec:simulation}) on personal devices, advance to a physical NeoRacer in classroom or club settings, and connect into college-level autonomous-racing programs. This pathway is co-designed with the RoboRacer Foundation~\cite{f1tenth2020} and complements the Boston University Autonomous Car Racing Organization (ACRO) model, whose operational structure is being codified as a startup kit for institutions seeking to launch student-led racing groups around the same hardware base.

\Paragraph{Post-Award Sustainability.}
The platform is designed to remain available as shared infrastructure independent of any single funding source. Hardware production runs through an established manufacturing relationship with a single fabrication partner. Software, simulation assets, and operating-system images follow a semester-aligned release cadence using the traceability scheme described in \S\ref{sec:traceability}. Kit revenue is reinvested into subsequent production batches and continued support, and institutional sponsorships fund the Sponsor-a-Kit program. This operational model mirrors other open-hardware foundations and is intended to keep the platform usable beyond the lifetime of any individual research grant.

\section{Licensing}
\label{sec:licensing}

All projects under the Neobotics Foundation are open-source, with the license chosen per repository to match the nature of the artifact. Hardware documentation is released under the CERN Open Hardware Licence Version 2, Strongly Reciprocal variant (CERN-OHL-S v2), which requires that derivative designs remain open and ensures that community improvements are shared back. To be explicit about what the current release contains: the published set includes the complete OSCORE electrical documentation (hardware manual, full schematic, and reference designator map), board interface descriptions and views, and 3D mechanical models of both the control board and the full vehicle assembly. PCB fabrication files (Gerbers) and editable EDA sources are not yet published, so the current release supports use, repair, interfacing, and mechanical integration rather than independent re-fabrication. The next NeoRacer hardware revision is being actively developed to be fully open-hardware, with complete fabrication sources included in the release. Software, including the ROS2 drivers, firmware, tools, and the student-facing \texttt{racecar-neo-library} (\S\ref{sec:stable_api}), is released under the GNU General Public License v3 (GPLv3), which applies the same reciprocal principle to code. Users should consult the \texttt{LICENSE} file in each repository for the specific terms applicable to that component.

\section{Conclusion}
\label{sec:conclusion}

We have presented NeoRacer, an open autonomous racing platform that addresses the cost and capability gaps in current robotics education and research infrastructure. By combining research-grade hardware (67~TOPS, LiDAR, 120~fps global shutter camera) with a pre-assembled, modular form factor, NeoRacer lowers the barrier to entry for autonomous vehicle education while providing the standardized benchmarking environment that the field currently lacks.

A pilot deployment at MIT (15 students, 2 weeks) validated the platform in an intensive educational setting and informed hardware revisions, including a camera upgrade from 30~fps to 120~fps and a power board redesign. These revisions demonstrate the iterative, deployment-driven development process that an open platform supports.

The spec-locked design, open licensing under CERN-OHL-S v2 and GPLv3, and compatibility with the F1Tenth ecosystem position the platform as a standardised foundation for cross-institutional algorithm benchmarking and for a scalable global competition ecosystem.

Hardware documentation, firmware, and ROS2 software packages are available at\\
{\small\url{https://github.com/Neobotics-Foundation-Inc/}}.

\section*{Acknowledgments}

The authors thank Dr.\ Sertac Karaman (MIT LIDS) and Dr.\ Renato Mancuso (BU CPS Lab) for advisory support; the 15 MIT IAP participants whose testing under competitive conditions drove critical hardware improvements; and Seeed Technology Co., Ltd.\ for manufacturing and co-development partnership.


\bibliographystyle{elsarticle-num}
\bibliography{references}

\end{document}